\documentclass{article}
\usepackage{spconf,amsmath,graphicx,hyperref}

\usepackage[dvipsnames]{xcolor}
\usepackage{booktabs}
\usepackage{amsfonts}
\usepackage{bm}
\usepackage{mathrsfs}

\title{BaLEEN: Biasing with Latent Encoded ENtities for context-aware ASR}

\name{Chihiro Taguchi$^{\star}$\sthanks{Work done at Sakana AI.} \qquad Yotaro Kubo$^{\dagger}$ \qquad Rujikorn Charakorn$^{\dagger}$}
\address{$^{\star}$ University of Notre Dame, Department of Computer Science and Engineering, IN, USA \\
         $^{\dagger}$ Sakana AI, Tokyo, Japan}

\begin{document}
\ninept
\maketitle
\begin{abstract}

Transcribing domain-specific entities and rare proper nouns remains a major challenge in automatic speech recognition (ASR).
In this paper, we propose \textbf{BaLEEN} (Biasing with Latent Encoded Entities), a lightweight, hypernetwork-based framework for dynamic contextual adaptation without fine-tuning the underlying ASR model.
BaLEEN encodes variable-length contextual keywords using a pretrained language model, compresses them into a fixed sequence of latent vectors via a Perceiver bottleneck, and injects context-dependent bias vectors directly into the intermediate encoder representations of the ASR model.
Because both the language model and the backbone ASR model remain entirely frozen during training, BaLEEN operates as a plug-and-play adapter that incurs zero computational overhead at inference time when context biases are precomputed.
We evaluate our method on a CTC-based ASR model using a Wikipedia-derived corpus with annotated named entities and synthetic speech.
Experimental results demonstrate that BaLEEN reduces keyword miss rate by 8.7\% on the test set relative to the unbiased baseline while simultaneously improving overall word error rate by 21\% and character error rate by 28\%.

\end{abstract}

\section{Introduction}
\label{sec:intro}

Context-awareness in automatic speech recognition (ASR) remains a central challenge in the field, even as ASR technology has been widely adopted in practical applications.
To achieve higher recognition accuracy and tailored user experiences, the model must transcribe speech by appropriately accounting for context in which they are deployed.
For example, in mobile ASR, the model should recognize proper nouns such as the contact names stored on the device.
Similarly, an ASR model for specific business domains should accurately identify specialized terminology and organization names.
Because the spelling of these named entities can be highly arbitrary and irregular, modern ASR architectures require access to external contextual information for transcribing them accurately.

Since many modern ASR models rely on attention mechanisms, several proposed extensions inject contextual phrases directly into a cross-attention module \cite{pundak2018deepcontextendtoendcontextual,kim2018dialog}.
Another common approach integrates an external language model to adjust token probability distributions \cite{zhao19d_interspeech}.
However, both types of the extensions incur additional computational cost and search latency at inference time, with cross-attention scaling linearly with the number of contextual keywords.

Alternatively, fine-tuning can adapt a model to a specific target domain if sufficient domain-specific audio--text pairs are available.
For instance, LoRA \cite{hu2022lora} is a widely adopted parameter-efficient method for domain adaptation that only trains lightweight adapter matrices while keeping the backbone model frozen.
However, acquiring sufficient data and training a dedicated model for every domain with arbitrary context poses severe practical limitations due to high costs.

In this paper, we propose \textbf{BaLEEN} (Biasing with Latent Encoded Entities), an adaptation method that relaxes training-data constraints without modifying or updating the target ASR model's parameters.
Our approach utilizes a hypernetwork architecture \cite{ha2017hypernetworks}, a neural network designed to generate parameters for another network.
Specifically, we construct a hypernetwork that converts context embeddings into layer-wise offset parameters for the ASR model.
By integrating these generated parameters directly into hidden representations, one can obtain a task-specialized model instantly, eliminating the need to fine-tune or retrain the backbone ASR architecture.

To train and evaluate hypernetwork-based biasing under rich domain diversity, we construct a synthetic dataset comprising 144k audio samples derived from English Wikipedia articles using LLM entity extraction and neural text-to-speech (TTS) synthesis.
Using this dataset, we train a Perceiver-based hypernetwork \cite{pmlr-v139-jaegle21a} to generate bias injected to the English ParakeetCTC baseline.
When contextual bias vectors are precomputed, our method incurs virtually zero computational overhead at inference time.
Experimental results demonstrate that BaLEEN reduces the keyword miss rate (KMR) by 14.3\% relative to the unbiased baseline while simultaneously improving overall word and character error rates.
Our code, dataset, and trained models will be made publicly available.

\section{Related work}
\label{sec:related}

Prior adaptation methods can be broadly categorized into three paradigms: decoding-time biasing, training a separate bias encoder module, and biasing the decoding CTC layer.

\noindent
\textbf{Decoding-time biasing.}
The earliest approaches leave the acoustic model untouched and intervene during decoding.
Shallow fusion interpolates an external LM into the beam search score computation \cite{zhao19d_interspeech}, and trie- or WFST-based deep biasing restricts and reweights hypotheses using a prefix structure built from a keyword list \cite{le21_interspeech}.
More recent work has shifted toward training-free setups, utilizing CTC-based word spotting to trigger boosting \cite{andrusenko24_interspeech} or synthetic multi-pronunciation tries to bias a frozen model zero-shot \cite{liu2025zeroshotcontextbiasingtriebased}.
For autoregressive models, prompt prefixing has also emerged as a practical decoding-time solution \cite{radford2022robustspeechrecognitionlargescale}.

\noindent
\textbf{Learned bias encoders.}
Another line of research directly models the interaction between acoustic features and contextual information.
CLAS introduced an attention-based bias encoder that embeds each phrase, allowing the decoder to attend to the resulting embedding set \cite{pundak2018deepcontextendtoendcontextual}.
The context-aware Transformer-Transducer extended this mechanism to multi-head attention over both audio and label representations \cite{chang2021contextawaretransformertransducerspeech}.
Contextual adapters made this framework parameter-efficient by inserting lightweight attention modules into a frozen transducer \cite{sathyendra2022contextualadapterspersonalizedspeech}, which was later refined using learned gating to deactivate biasing on non-entity frames \cite{Alexandridis2023}.
However, because these approaches represent the bias list as one embedding per phrase, the computational cost of attending to the list scales linearly with its size, and the attention distributions become increasingly diffuse on large lists.
The neural associative memory line makes this tension explicit, addressing it with two-pass top-$K$ retrieval so that the acoustic encoder attends only to a shortlist \cite{9747726,10023323,wu23e_interspeech,wu2024deferrednamlowlatencytopk}.

\noindent
\textbf{Biasing CTC models.}
Because CTC decoding is non-autoregressive and assumes frame-level conditional independence, contextual information must be injected into the encoder
representations or frame posteriors rather than a decoder state.
CPPNet fuses an attention-derived context vector into the encoder output and
adds an auxiliary contextual-phrase prediction loss
\cite{huang23d_interspeech}, whereas Zhang et al. \cite{zhang2022endtoendcontextualasrbased} add an attention-weighted bias score
directly to the CTC linear projections in a hybrid CTC/attention architecture.
Other approaches inject contextual supervision at intermediate encoder layers
\cite{Shakeel_2024,nakagome24_interspeech,nakagome25_interspeech}, apply biasing exclusively at CTC spike frames \cite{huang2023spiketriggered}, or expand the output
vocabulary with dedicated bias tokens \cite{lin25g_interspeech}.

Unlike these methods incurring additional computation to inject keyword bias, our proposed \textbf{BaLEEN} uses a hypernetwork to add latent contextual bias vectors directly into the backbone model without changing its decoding pipeline.

\begin{figure}[t]
    \centering
    \includegraphics[width=1\linewidth]{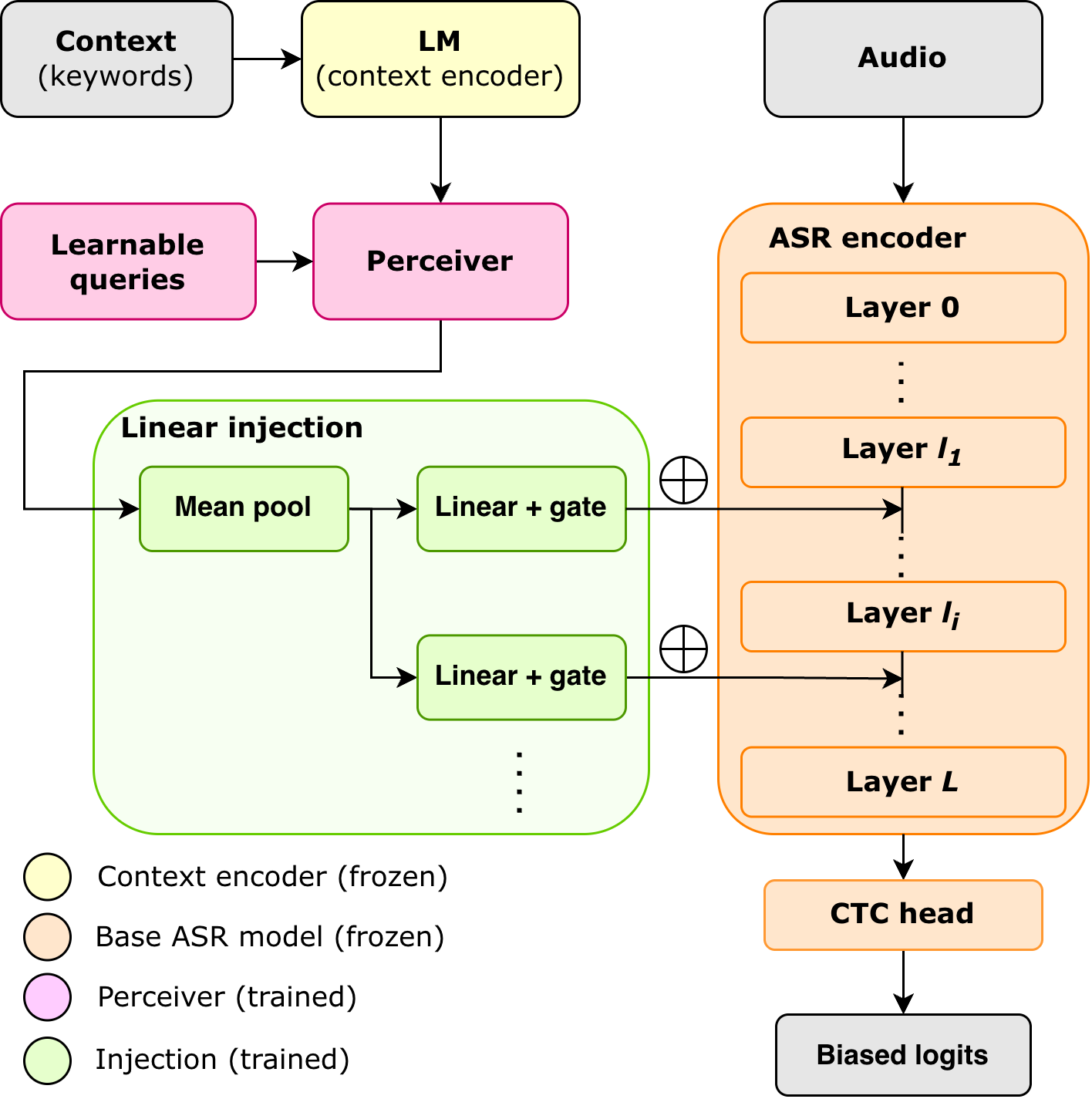}
    \caption{Our proposed methods for context biasing.
    The linear injection layer is instantiated for each target ASR layer.}
    \label{fig:contextbiasing}
\end{figure}

\begin{figure*}[t]
    \centering
    \includegraphics[width=.75\linewidth]{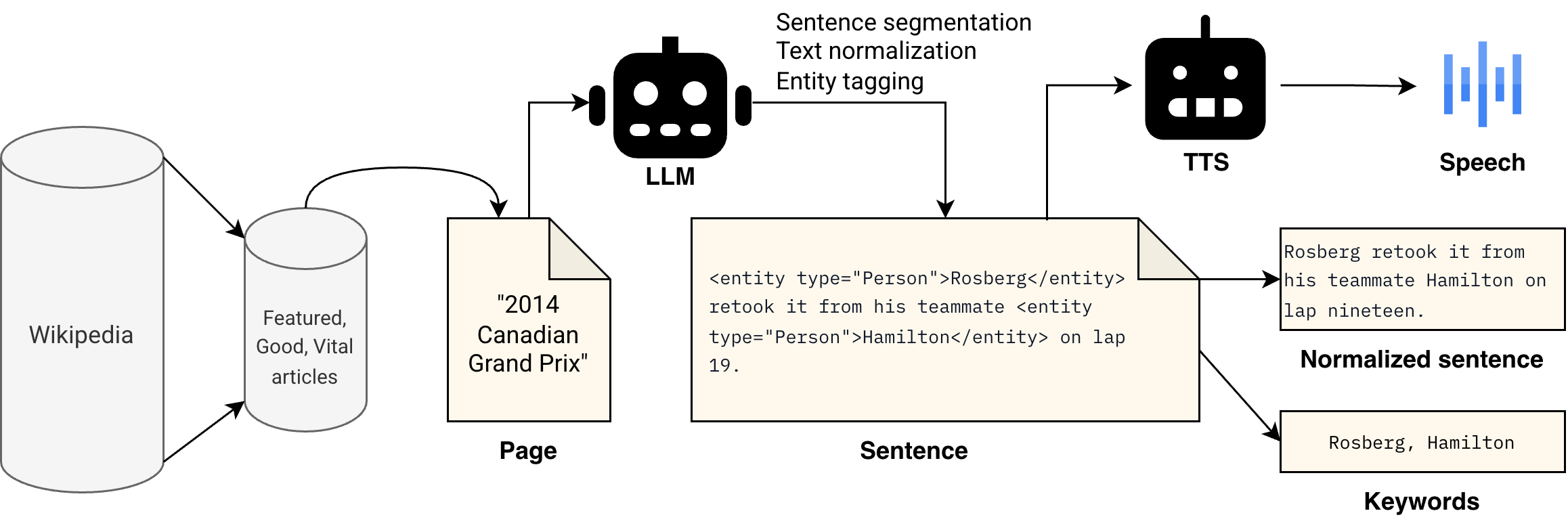}
    \caption{Synthetic dataset construction flow.}
    \label{fig:dataset_creation}
\end{figure*}

\section{Method}\label{sec:method}

Our architecture comprises three main components: (1) a language model (LM) for context encoding, (2) a CTC-based ASR backbone, and (3) a bottleneck module that compresses the encoded context and injects contextual knowledge into the ASR model.
To convert a conventional ASR model into a dynamically adaptable system, we adopt a hypernetwork framework that computes context-dependent bias parameters for the ASR model.
Figure~\ref{fig:contextbiasing} provides a schematic overview of our method.

In this architecture, the model is designed to handle arbitrary textual context.
The LM converts the context into a sequence of representation vectors, and the context bottleneck compresses these token-wise representations into fixed-size semantic representations.
Because contextual inputs vary in length, we implement the bottleneck using a Perceiver network \cite{pmlr-v139-jaegle21a}.
The Perceiver compresses variable-length sequence inputs into a fixed-length latent sequence via cross-attention with learnable query vectors.
While originally proposed for non-text modality integration (\textit{e.g.}, vision), the Perceiver has also proven effective for compressing text \cite{charakorn2024instant}.

Formally, let $x$ denote the input context.
We extract hidden outputs from the $\ell$-th intermediate layer of the LM and project them into $d$-dimensional embeddings: $C = \operatorname{FFN}(\operatorname{LM}^{(\ell)}(x)) \in \mathbb{R}^{n \times d}$, where $n$ is the number of context tokens.
Let $Q \in \mathbb{R}^{N \times d}$ denote the learnable latent query vectors, where $N$ and $d$ are the latent query length and hidden dimension, respectively (set to $N = 16$ and $d = 512$ by default).\footnote{For simplicity, normalization layers are omitted in equations henceforth; see our implementation details released publicly upon acceptance.}.
The sequence $C$ passes through a single Perceiver block consisting of cross-attention followed by self-attention, with residual connections around each layer: $M = \operatorname{Perceiver}(Q, C) \in \mathbb{R}^{N \times d}$.

The compressed representation $M$ is subsequently injected into targeted intermediate layers of the ASR encoder via linear projections.
For a selected subset of target ASR layers $\mathcal{L} \subseteq \{1, \dots, L\}$ (where $L$ is the total number of encoder layers), $M$ is mean-pooled across its sequence dimension into a vector $m = \pi(M) \in \mathbb{R}^{d}$.
For each target layer $l \in \mathcal{L}$, $m$ is projected through a layer-specific weight matrix $W_l \in \mathbb{R}^{d \times d_\text{ASR}}$ and bias vector $b_l \in \mathbb{R}^{d_\text{ASR}}$:
\begin{equation}
    B_l = W_l^\top m + b_l \in \mathbb{R}^{d_\text{ASR}}.
\end{equation}
The resulting contextual bias vector $B_l$ is injected into the hidden representation matrix $H^{(l)} \in \mathbb{R}^{T \times d_\text{ASR}}$ across all $T$ acoustic frames:
\begin{equation}
    H^{(l)} \leftarrow H^{(l)} + \gamma_l \bm{1}_T B_l^\top,
\end{equation}
where $\bm{1}_T \in \mathbb{R}^{T \times 1}$ is a column vector of ones and $\gamma_l \in \mathbb{R}$ is a scalar scaling parameter.
At the beginning of training, $\gamma_l$ is initialized to zero for all $l$ to avoid catastrophic forgetting.
Under this design, the generated context bias is independent of the audio input and can be added to the ASR intermediate representations in a modular, plug-and-play fashion.
The user can easily disable contextual biasing at inference time by detaching the hypernetwork module or setting $\gamma_l = 0$.

A natural question regarding uniform bias injection is whether adding a static, time-invariant offset $B_l$ across all time frames risks degrading non-keyword speech or triggering false positives.
Our framework is motivated by the hypothesis that the internal self-attention and non-linear projections of the backbone ASR encoder serve as dynamic temporal filters.
While $B_l$ remains constant over time, local acoustic representations vary.
When acoustic evidence phonetically aligns with a target entity, self-attention and feed-forward projections amplify the context-biased manifold.
Conversely, on frames lacking relevant acoustic support, non-linear activations suppress $B_l$, mapping hidden states back toward standard subword or CTC blank ($\epsilon$) posterior distributions.
By distributing this static injection across all encoder layers, the network achieves progressive, multi-stage contextual steering without requiring frame-wise cross-attention or on-the-fly context encoding at inference time.

The model is trained by minimizing the standard CTC loss.
Throughout training, the parameters of both the context encoder LM and the backbone ASR model remain frozen, and only the parameters of the auxiliary context bottleneck and projection layers are updated.
At inference time, when the context set is fixed, the bias term $\gamma_{l} \bm{1}_T B_l^\top$ can be precomputed once.
As a result, the bias acts as a set of static offset parameters, avoiding any recomputation through the LM or bottleneck module and incurring virtually zero runtime overhead.

\section{Dataset}
\label{sec:dataset}

Training the context bottleneck requires triplet data consisting of (audio, transcription, keyword list) across a sufficiently diverse sample space to achieve robust generalized context adaptation.
To address this, we construct a synthetic dataset comprising 144k audio samples derived from 993 English Wikipedia articles using LLM-powered entity tagging and text-to-speech (TTS) synthesis.
Figure~\ref{fig:dataset_creation} illustrates the dataset generation pipeline to obtain the triplet data.

We leverage Wikipedia to construct a corpus spanning diverse domains with high keyword density and high-quality text.
First, we extract the Featured Articles\footnote{\url{https://en.wikipedia.org/wiki/Wikipedia:Featured_articles}}, Good Articles\footnote{\url{https://en.wikipedia.org/wiki/Wikipedia:Good_articles}}, and Vital Articles\footnote{\url{https://en.wikipedia.org/wiki/Wikipedia:Vital_articles}}, curated by Wikipedia editors based on completeness, neutrality, and prose depth, using the Wikipedia-API Python library\footnote{\url{https://pypi.org/project/Wikipedia-API}}.
We then filter out disambiguation pages and articles shorter than 3,000 characters.
Next, each remaining article is segmented into sentences and annotated with named entity tags using GPT-5.4 Mini.
Finally, each sentence is converted to speech using Gemini 3.1 Flash TTS, where a voice is randomly sampled from 30 prebuilt speaker profiles to ensure acoustic and speaker diversity.

The dataset supports two keyword sampling configurations: (1) using only the keywords that appear strictly in the spoken audio transcription, or (2) using all keywords appearing within the source article.
The latter configuration introduces contextually relevant hard-negative distractors alongside spoken keywords, creating a more challenging training scenario that fosters stronger context adaptability.

We partition the dataset into 80\% training, 10\% validation, and 10\% test splits.
Crucially, these splits have zero overlap in their source articles (\textit{i.e.}, unique contexts) to prevent context leakage during evaluation.

\section{Experiments}

To encode keywords in a spelling-aware manner, we extract hidden representations from layer $\ell$ ($\ell = 7$ by default) of ByT5 \cite{xue-etal-2022-byt5}.
For the ASR backbone, we employ the CTC variant of parakeet-tdt\_ctc-110m \cite{rekesh2023fastconformer}, which consists of 17 Conformer encoder layers followed by a CTC layer trained on English.
The CTC vocabulary comprises 1,025 case-sensitive subword tokens constructed via Byte-Pair Encoding \cite{sennrich-etal-2016-neural}.
To prevent out-of-vocabulary (OOV) issues arising from the CTC model's relatively small subword vocabulary, all transcript text in the dataset is normalized.
For example, because the ParakeetCTC vocabulary does not contain numeric digits, numbers are expanded into spelled-out words (e.g., ``3'' to ``three'') by the sentence processing LLM at the dataset creation phase (cf.~Figure~\ref{fig:dataset_creation}).
Additionally, non-phonetic punctuation marks (excluding periods and commas) are removed prior to TTS synthesis.
Utterances containing any remaining unsupported characters are filtered out from both training and evaluation splits.
To evaluate performance, we report the \emph{keyword miss rate} (KMR), which is defined as the proportion of ground-truth keywords omitted in predictions ($1 - \text{Recall}$), alongside word error rate (WER) and character error rate (CER).

To investigate which ASR layers are critical for keyword biasing, we evaluate four distinct layer selection configuration s$\mathcal{L}$: (1) later layers ($\mathcal{L} = \{l^{(13)}, l^{(14)}, l^{(15)}, l^{(16)}\}$), (2) middle-to-later layers ($\mathcal{L} = \{l^{(10)}, l^{(12)}, l^{(14)}, l^{(16)}\}$), (3) uniformly distributed layers ($\mathcal{L} = \{l^{(4)}, l^{(8)}, l^{(12)}, l^{(16)}\}$), (4) all layers ($\mathcal{L} = \{l^{(i)}\}_{i=0}^{16}$).
Additionally, we evaluate whether augmenting training context lists with probabilistically sampled hard negatives from the same source article enhances model generalization.
Note that dynamic distractor sampling is used only during training, while evaluation contexts remain fixed with keywords strictly appearing in the utterance.
In total, $4 \times 2 = 8$ settings are compared against the non-fine-tuned ASR baseline.
The resulting context bottleneck module comprises 37.7M parameters, with each linear layer projection adding 2.6M parameters.

Unless specified otherwise, we adopt the following hyperparameter settings across all experiments: a learning rate of $5 \times 10^{-5}$ with a 10\% linear warmup followed by linear decay, 10 training epochs, a batch size of 32, and a single Perceiver encoder block.
The Perceiver decoder was omitted after preliminary experiments indicated it provided no measurable performance benefit.
For each sample, keywords were concatenated into a single string, separated by a comma and a space.
For training stability and batching efficiency, speech samples shorter than 1 second or longer than 20 seconds are excluded from training.
All experiments are run on an NVIDIA H100 GPU (80GB), and all reported metrics are computed using greedy decoding.

Inspired by prior work using cross-attention for contextual injection \cite{chang2021contextawaretransformertransducerspeech} we also implement a direct-biasing baseline that modifies logits $\bm{s} = \operatorname{CTC}(H^{(L)})$ by integrating the compressed vectors $M$ through cross-attention: $\bm{s} \leftarrow \bm{s} + \operatorname{XAttn}(H^{(L)}, M, M)$.

\section{Results}
\label{sec:results}

\begin{table}[t]
    \centering
    \setlength{\tabcolsep}{3.8pt}
    \begin{tabular}{@{}lrrr@{}} \toprule
        Architecture & WER$^{(\downarrow)}$ & CER$^{(\downarrow)}$ & KMR$^{(\downarrow)}$ \\ \midrule
        Baseline (no FT) & 9.53 & 2.74 & 40.66 \\ 
        Baseline (XAttn + logits) & 11.47 & 3.29 & 35.59 \\
        Baseline (XAttn + logits), DC & 11.16 & 2.58 & 35.26 \\ \midrule
        Linear + hidden (late) & 7.97 & 2.06 & 36.37 \\
        Linear + hidden (midlate) & 7.86 & 2.06 & 36.15 \\
        Linear + hidden (uniform) & 7.86 & 2.04 & 35.67 \\
        Linear + hidden (all) & 7.62 & 1.98 & 35.71 \\ \midrule
        Linear + hidden (late), DC & 7.75 & 2.01 & 35.57 \\
        Linear + hidden (midlate), DC & 7.97 & 2.09 & 35.81 \\
        Linear + hidden (uniform), DC & 8.31 & 2.12 & 35.47 \\
        Linear + hidden (all), DC & \textbf{7.38} & \textbf{1.87} & \textbf{34.85} \\
        \bottomrule
    \end{tabular}
    \caption{Evaluation results on the validation set.
    All metrics are reported in percentages (\%).
    DC stands for dynamic context with probabilistically mixed hard-negative keywords.
    }
    \label{tab:main-results}
\end{table}

Table~\ref{tab:main-results} reports the performance metrics for model checkpoints achieving the lowest KMR on the validation set.
All proposed adaptation variants improve keyword recognizability while simultaneously outperforming the non-fine-tuned baseline ASR model in overall WER and CER.
Incorporating hard-negative distractors into the contextual keyword list during training further enhances model robustness.
The overall best-performing configuration, injecting linearly generated biases into every ASR encoder layer using dynamic context with hard-negative distractors, reduces WER by 22.6\%, CER by 31.7\%, and KMR by 14.3\% relative to the baseline.
While directly biasing output logits via cross-attention also improves keyword recognition, it exhibits training instability, consistently appending an extraneous random token to the end of predicted transcripts, likely due to unconditioned cross-attention corrupting the CTC blank token distribution.

Evaluation on the test set (Table~\ref{tab:test-results}) reveals similar trends.
Direct logit biasing via cross-attention without distractor keywords achieves the lowest KMR, yielding an 11.0\% relative improvement over the non-fine-tuned baseline.
Nevertheless, it suffers from the same decoding instability, consistently appending spurious tokens and degrading overall WER and CER.
Furthermore, this KMR gain occurs only when the keyword lists are restricted to keywords uttered in the audio; however, this idealized setup is rarely encountered in real-world deployments.
Under realistic dynamic context conditions, hidden-state biasing across all ASR layers yields the best balanced performance.
This setting reduces WER by 21\% and CER by 28\% relative to the non-fine-tuned baseline, while maintaining a competitive 8.7\% relative reduction in KMR.
It is also important to note that the compressed context vectors in the direct logit biasing baseline can attend to encoded acoustic frames with incurred computational costs.
In contrast, our proposed method injects bias under a more restrictive condition agnostic of the audio input.

Comparing late, middle-late, and uniform 4-layer injection strategies indicates that among partial layer subsets, the specific choice of targeted encoder layers does not significantly alter performance.
However, comparing performance metrics across the validation and test sets reveals that targeting only a subset of ASR encoder layers yields weak generalization, while biasing all layers with dynamic context simultaneously shows improvement over the non-fine-tuned baseline.

Finally, constructing a large-scale synthetic dataset was necessary to evaluate contextual biasing under dense, arbitrary keyword lists.
Accordingly, the primary objective of our experiments was to establish context generalizability under these complex synthetic conditions, which was validated by our results.
Nevertheless, we acknowledge that relying on synthesized speech presents a practical limitation regarding generalization to real-world human voices, accents, and diverse acoustic environments.

\begin{table}[t]
    \centering
    \setlength{\tabcolsep}{3.8pt}
    \begin{tabular}{@{}lrrr@{}} \toprule
        Architecture & WER$^{(\downarrow)}$ & CER$^{(\downarrow)}$ & KMR$^{(\downarrow)}$ \\ \midrule
        Baseline (no FT) & 10.36 & 3.04 & 43.34 \\
        Baseline (XAttn + logits) & 11.19 & 3.21 & \textbf{38.59} \\
        Baseline (XAttn + logits), DC & 13.91 & 5.12 & 42.32 \\ \midrule
        Linear + hidden (late) & 9.51 & 2.63 & 45.92 \\
        Linear + hidden (midlate) & 9.16 & 2.49 & 44.41 \\
        Linear + hidden (uniform) & 8.70 & 2.36 & 41.32 \\
        Linear + hidden (all) & 9.22 & 2.56 & 44.05 \\ \midrule
        Linear + hidden (late), DC & 8.69 & 2.36 & 42.24 \\
        Linear + hidden (midlate), DC & 9.78 & 2.72 & 47.39 \\
        Linear + hidden (uniform), DC & 10.05 & 2.70 & 43.48 \\
        Linear + hidden (all), DC & \textbf{8.18} & \textbf{2.19} & 39.55 \\
        \bottomrule
    \end{tabular}
    \caption{Evaluation results on the test set.
    All metrics are reported in percentages (\%).
    }
    \label{tab:test-results}
\end{table}

\section{Conclusion}
\label{sec:conclusion}

This paper introduced \textbf{BaLEEN} (Biasing with Latent Encoded Entities), a lightweight, plug-and-play contextual biasing framework based on hypernetworks.
Our method encodes and compresses arbitrary, variable-length contextual entities into fixed-length latent vectors, which are then injected into intermediate hidden representations of a frozen ASR model.
This approach enables context-aware adaptation without modifying or fine-tuning the underlying backbone weights.
Furthermore, because the generated bias vectors can be precomputed for a static context list, runtime computational overhead and latency are zero during inference.
To train and benchmark our model, we constructed a large-scale Wikipedia-based synthetic dataset consisting of triplets of audio, transcripts, and contextual keyword lists.
Experimental evaluations demonstrate that injecting linearly projected bias vectors across all ASR encoder layers achieves the best overall performance, reducing the KMR by 14.3\% while simultaneously improving WER and CER over the baseline.
In addition, we showed that incorporating hard-negative distractors into the context list during training enhances model generalization under realistic deployment conditions.

\vfill\pagebreak

\newlength{\bibitemsep}
\setlength{\bibitemsep}{0.05em}
\let\oldthebibliography\thebibliography
\renewcommand\thebibliography[1]{%
  \oldthebibliography{#1}%
  \setlength{\itemsep}{\bibitemsep}%
}
\bibliographystyle{IEEEbib}
\bibliography{refs}

\end{document}